\documentclass[sigplan,nonacm]{acmart}
\newcommand{\arch}[1]{Agentic Architect}

\newcommand*\circled[1]{\raisebox{.5pt}{\textcircled{\raisebox{-.9pt} {#1}}}}
\usepackage{listings}
\usepackage{multirow}
\usepackage{tabularx}
\usepackage{xfrac}
\usepackage{tikzducks}

\begin{document}
\title{\arch{}: An Agentic AI Framework for Architecture Design Exploration and Optimization}
\author{Vasilis Kypriotis}
\affiliation{%
  \institution{Carnegie Mellon University}
  \city{Pittsburgh}
  \state{PA}
  \country{USA}
}
\author{Alexander Blasberg}
\affiliation{%
  \institution{Carnegie Mellon University}
  \city{Pittsburgh}
  \state{PA}
  \country{USA}
}
\author{Alexandros Moiras}
\affiliation{%
  \institution{Carnegie Mellon University}
  \city{Pittsburgh}
  \state{PA}
  \country{USA}
}
\author{Dimitrios Skarlatos}
\affiliation{%
  \institution{Carnegie Mellon University}
  \city{Pittsburgh}
  \state{PA}
  \country{USA}
}

\begin{abstract}

Computer Architecture research has traditionally required the exploration of vast, highly complex design spaces through manual human effort. Large Language Models (LLMs) offer an opportunity to accelerate this process by enabling fast and automated implementation and testing of different architectural mechanisms and their combinations.

We introduce \arch{}, a framework for agentic CPU microarchitectural design-space exploration and optimization. \arch{} combines LLM-driven code evolution with cycle-accurate simulation and hardware cost modeling to explore performance, area, and power tradeoffs. Starting from a seed design, the LLM is prompted to explore the design space by implementing changes, guided by prior evaluation feedback. Each new design is compiled, simulated on training workloads, evaluated against the optimization objective, and constrained by area and power budgets. Its score and diagnostics update the candidate population and guide subsequent iterations.

We evaluate \arch{} across prefetching, cache replacement, and branch prediction. On 213 traces spanning five benchmark suites, its evolved prefetchers lead the evaluated L1D and L2 designs, delivering up to 6.7\% and 2.6\% additional speedup, respectively, over the strongest state-of-the-art design, while the best evolved cache replacement design improves over Mockingjay by up to 1.7\%. Moreover, across 191 branch-prediction traces, the evolved branch predictor achieves 97.62\% of TAGE-SC-L's geometric mean IPC at 10.58\% of its area. Our analysis highlights the importance of seed selection, hardware constraints, and training workloads in shaping search outcomes. Overall, \arch{} establishes the first end-to-end framework for agentic CPU microarchitectural design optimization, opening a new direction for architecture research. The framework will be open sourced upon publication.

\end{abstract}




\maketitle % should come after the abstract
 
\section{Introduction}
\label{sec:intro}

With the slowdown of Moore's Law and the rising cost of larger on-chip structures, improving processor performance and efficiency increasingly depends on architectural design-space exploration. Decisions about mechanisms, interacting signals, control logic, and hardware organization create a vast combinatorial space. Architects traditionally explore this space by proposing a design, implementing it, simulating it, inspecting the results, and iterating. Each design must be evaluated for performance across workloads and respect area, power, and timing constraints. The effort required limits how many implementations architects can examine, leaving potential combinations and refinements unexplored.

These challenges make a case for an agentic approach that automates and accelerates the architectural design loop. Recent agentic AI systems pair LLMs with evolutionary search~\cite{alphaevolve,funsearch,openevolve,adaevolve,codeevolve,shinkaevolve,deepevolve,thetaevolve}, demonstrating the potential for AI-driven discovery across scientific and engineering domains. Computer architecture is a compelling but demanding target for this approach. Designs are naturally expressed as modular code, cycle-accurate simulators enable consistent evaluation, and the underlying design space is large but well-structured. However, exploring architectural design remains difficult because performance is workload-dependent, hardware constraints limit feasible designs, and performance depends on multiple interacting mechanisms and signals. The central question of this paper is whether LLMs can generate architectural designs competitive with the state-of-the-art (SoTA), and uncover further gains in performance and efficiency.

To explore that, we introduce \arch{}, an agentic framework for CPU microarchitectural design-space exploration and optimization. It connects AI-driven code evolution with cycle-accurate simulation and hardware cost modeling to generate, evaluate, and refine candidate designs under a configurable objective. The framework is initialized with a seed design, LLM, evolutionary framework, architectural simulator, hardware model, task-specific prompt, training workloads, and hardware budgets, which together define the search space and evaluation conditions.

The evolutionary loop starts from the seed population. At each iteration, the search selects a design, combines its code with prior feedback, and prompts the LLM to produce a modified candidate design. The design is validated and evaluated for performance, area, and power. Candidates that satisfy the design constraints are scored against the optimization objective. These scores guide population updates and subsequent selection, while diagnostics inform further changes. After the iteration budget is exhausted, the highest-scoring design is evaluated on the full benchmark suite.

The framework's modular interfaces allow LLMs, evolutionary frameworks, simulators, and hardware models to be changed independently. Architects can therefore select tools appropriate to each architectural domain and configure the seed, optimization objective, and hardware constraints. This makes \arch{} a co-design framework: the architect establishes the search conditions, while the AI agent explores within them.

We instantiate \arch{} in data prefetching at both L1D and L2 caches, cache replacement and branch prediction, and we examine the factors that shape the evolution. First, seed selection presents a tradeoff between refining a strong design and exploring from a simpler seed. Both approaches prove useful in our experiments: the mature VA/AMPM prefetcher~\cite{vaampm} provided the foundation for the best evolved L1D design, while a basic stride prefetcher led to the best evolved L2 design. In replacement, even a simple LRU seed evolved into a competitive design. 

Hardware budgets are equally important: without area and power constraints, evolution can gain performance simply by expanding hardware state. Being bounded, the evolved designs improve over strong baselines with modest hardware cost, including only 3.7\% additional L1D area over MLOP~\cite{mlop} for the evolved L1D prefetcher, and 0.4\% additional LLC area over Mockingjay~\cite{mockingjay} for the evolved replacement design.

Finally, training-set composition is critical for evolving designs that generalize. 
Therefore, we selected training sets spanning diverse cache-miss or branch-misprediction rates, to cover a wider range of workload behaviour. Comparing speedups across the training traces and held-out traces shows that the evolved designs retain their performance advantages on unseen workloads.

Observing the evolved designs reveals interesting architectural insights. First, evolution tends to combine well-established mechanisms and signals into coordinated designs. Both evolved prefetchers combine several engines with local control or a central cost gate. Across the designs, it combines established signals, such as memory-level parallelism (MLP) and prefetch lateness, with new derived signals, such as cross-instruction reuse and relative memory-bank traffic. This is the behavior we sought to enable: exploring combinations whose manual refinement would require extensive experimentation. Under fixed hardware budgets, the challenge is not simply adding more state, but finding combinations retaining their benefits while sharing limited resources.

Finally, to assess the performance of evolved designs, we evaluate the evolved prefetching and replacement designs across 213 traces spanning SPEC CPU2006~\cite{spec2006}, SPEC CPU2017~\cite{spec2017}, PARSEC~\cite{parsec}, Ligra~\cite{ligra} and Google Workload Traces v2~\cite{google}. We report geometric mean Instructions-Per-Cycle (IPC) speedup on a single core and geometric mean weighted speedup on four cores.

In prefetching, the evolved L1D design achieves speedups of $1.349\times$ over no prefetching on a single core and $1.356\times$ on four cores. These provide an additional 4.9\% and 6.7\% speedup over MLOP~\cite{mlop}, the strongest evaluated published L1D design, and 7.8\% and 8.4\% over its VA/AMPM seed~\cite{vaampm}. At L2, the evolved prefetcher achieves speedups of $1.434\times$ and $1.420\times$, outperforming all evaluated L2 prefetchers, including SPP+PPF~\cite{ppf}, the strongest published design on our workloads, by 2.6\% and 2.0\%. 

In cache replacement, the best evolved design reaches $1.042\times$ and $1.059\times$ speedup over LRU on one and four cores, improving on Mockingjay by 1.4\% and 1.7\%, respectively. Starting from LRU, evolution discovers an MLP-based cost design that ranks lines by expected miss cycles saved rather than Mockingjay's Estimated Time of Reuse (ETR). Its better balance across workloads, particularly on unseen held-out traces, allows it to outperform Mockingjay overall.

Lastly, we evaluate branch prediction across 191 traces from the five benchmark suites, using IPC per watt as the optimization objective to explore tradeoffs beyond raw performance. The evolved predictor retains 97.62\% of TAGE-SC-L's~\cite{tagescl} geometric mean IPC, while reducing modeled area and power by 89.42\% and 4.62\%, respectively. This demonstrates that agentic search can identify efficient designs under objectives that balance performance, area, and power.

In summary, the contributions of this paper are:
\begin{itemize}
    \item \arch{}, the first framework for agentic CPU microarchitectural optimization, built on a modular architecture for evolutionary frameworks, LLMs, cycle-accurate simulators, and hardware modeling.
    \item A systematic analysis of how seed selection, hardware budgets, optimization objectives, and training workloads shape evolution, including evaluation on held-out workloads to assess generalization.
    \item An analysis of what evolutionary search can discover: combinations of established mechanisms and signals, and more efficient organizations under objectives beyond raw performance.
    \item Evolved designs that optimize performance or efficiency across three domains: cache replacement, data prefetching and branch prediction.
    \item Open-source release: The Agentic Architect framework will be open-sourced following publication.
\end{itemize}
\section{Background}
\label{sec:background}

\arch{} combines LLM-driven code evolution with cycle-accurate simulation for computer architecture design exploration and optimization. This section discusses the three domains and the evolutionary framework we build on.

\subsection{Microarchitecture Design}

\noindent\textbf{Cache replacement.} LLC replacement decides which lines to evict, how to prioritize incoming lines, and whether to bypass the cache. These decisions depend on expected reuse: LRU tracks recency, DRRIP~\cite{drrip} adapts insertion priorities, and PC-based predictors learn reuse from program context~\cite{ship,shippp}. Hawkeye~\cite{hawkeye} learns cache-friendly labels from Belady's optimal policy, while Mockingjay~\cite{mockingjay} predicts reuse distances to rank lines. Miss cost also matters: memory-level parallelism (MLP) overlaps requests, so retaining a line whose miss would stall execution can be more valuable~\cite{qureshi_mlp}.

\noindent\textbf{Hardware Prefetching.} Hardware prefetchers issue requests ahead of demand to hide latency and increase MLP. Common designs use prediction engines that capture complementary access patterns: 
next-line and PC-stride target regular accesses, spatial prefetchers use access maps or footprints~\cite{vaampm,sms,bingo}, delta predictors such as Berti~\cite{berti} and SPP~\cite{spp} learn address differences, and MLOP~\cite{mlop} scores offsets at multiple lookahead depths. These mechanisms predict both which data to fetch and how far ahead.

Combining prediction engines requires coordinating request issuance and fill level. Coverage, accuracy, and timeliness capture the resulting tradeoffs, as greater aggressiveness can increase coverage at the cost of bandwidth and cache pollution. Prefetchers therefore use filtering and adaptive feedback to balance these tradeoffs: PPF~\cite{ppf} rejects predictions likely to be unused, DSPatch~\cite{dspatch} adjusts coverage and accuracy according to bandwidth, and reinforcement-learning prefetchers such as Pythia~\cite{pythia} learn actions from rewards for usefulness, timeliness, and bandwidth conditions.

\noindent\textbf{Branch Prediction.} Branch predictors use history to predict branch outcomes and avoid costly mispredictions. Perceptron-based designs~\cite{perceptron,hashed_perceptron,mpp} combine learned weights from history features, while TAGE~\cite{tage} selects predictions from tagged tables using different history lengths. TAGE-SC-L~\cite{tagescl} adds a statistical corrector and loop predictor, forming a strong baseline for exploring performance and hardware efficiency.

\subsection{LLM-Driven Code Evolution}

\arch{} performs architectural exploration through evolutionary search, using LLMs to generate candidate design mutations. AlphaEvolve~\cite{alphaevolve} demonstrated this approach, while OpenEvolve~\cite{openevolve} provides an open-source framework following the same ideas. Its evolutionary loop has four stages: the \emph{LLM Evolver} mutates a selected implementation; the \emph{Evaluator} compiles and benchmarks the candidate; the \emph{Solution Database} stores candidates in islands to maintain diversity; and \emph{Prompt Creation} selects a parent and constructs the next prompt using evaluation feedback. Further works adapt island selection and exploration to search progress~\cite{adaevolve}, or improve efficiency through parent sampling, novelty filtering, and bandit-based LLM selection~\cite{shinkaevolve}.

Computer architecture suits this approach: designs are modular logic blocks that LLMs can modify, while cycle-accurate simulators and benchmarks provide an established evaluation loop. However, architectural evolution must further consider how mechanisms and signals interact with the core and memory hierarchy, alongside their hardware costs. \arch{} addresses these requirements by enforcing modeled area and power budgets and guiding code changes with feedback on architectural behavior, according to the architectural domain. We discuss these challenges through the design and evaluation of \arch{}.
\section{\arch{}'s Design}
\label{sec:design}

\begin{figure*}[t]
\centering
\includegraphics[width=\textwidth]{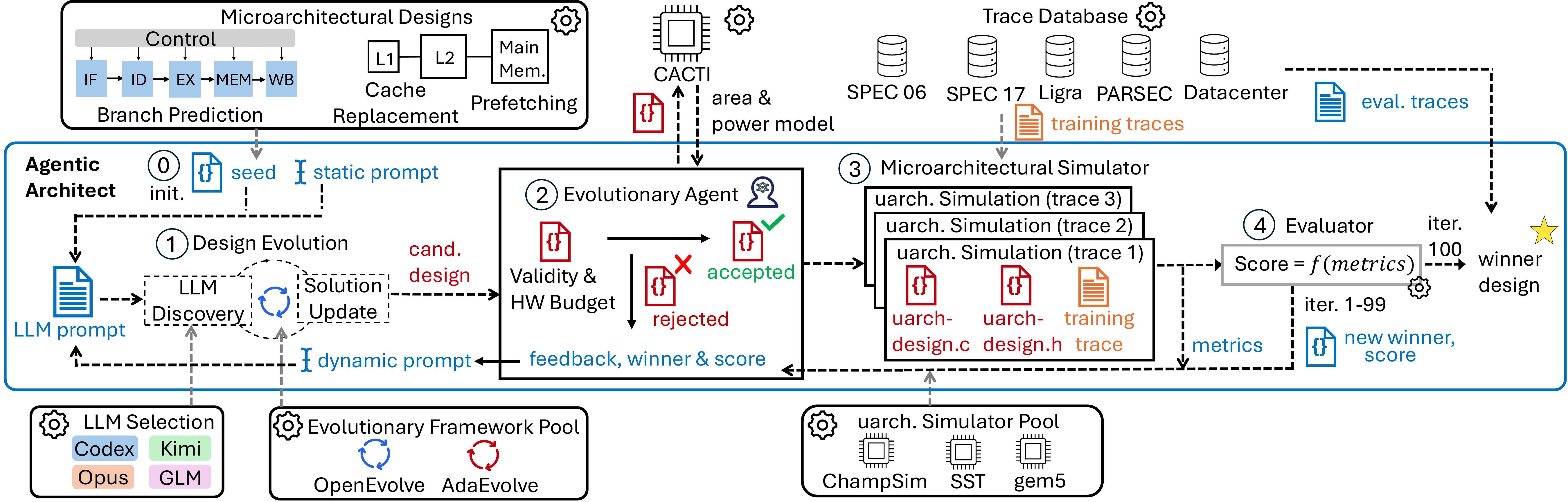}
\caption{Overview of the \arch{} framework. Steps~\circled{0}--\circled{4} show initialization, design evolution, candidate validation, microarchitectural simulation, and evaluation. Gear icons mark configurable framework components.}
\label{fig:design}
\end{figure*}

The \arch{} framework automates microarchitectural design-space exploration as a constrained optimization process. It generates, evaluates, and refines candidate designs toward an architect-defined objective, such as IPC or performance per watt, while enforcing hardware budgets throughout the search. 

Figure~\ref{fig:design} illustrates the end-to-end workflow. Evolution is initialized at Step~\circled{0} by a baseline design called \textit{the seed} and a static prompt, which provides initial guidance for the search. The remaining inputs, marked by gears in the figure, configure the framework: the LLM and evolutionary framework, simulator and hardware model, trace database, optimization objective, and hardware budgets. Together, these configurations remain fixed throughout the evolutionary loop.

At each iteration, design evolution (Step~\circled{1}) produces candidate designs using the chosen LLM and evolutionary framework. Each generated candidate is passed to the evolutionary agent (Step~\circled{2}), which checks its validity and hardware feasibility. Accepted candidates are executed on the training traces by the microarchitectural simulator (Step~\circled{3}), which returns execution metrics to the evaluator (Step~\circled{4}). The evaluator converts these measurements into a score to guide subsequent iterations. Once the iteration budget is exhausted, the highest-scoring design is evaluated on the held-out traces. The remainder of this section describes the framework by following the evolutionary workflow.

\subsection{Initialization and Configuration}

Architectural design tasks differ in evaluation tools, objectives, constraints and more. For instance, tasks may use cache misses, prefetch coverage, or branch mispredictions as metrics, permit changes to mechanisms or storage organization, and optimize IPC or performance per watt under different hardware budgets. \arch{} separates these choices into \textit{evolution initialization}, which defines the starting point for search, and \textit{framework configuration}, which determines how candidates are generated and evaluated.

Initialization (Figure~\ref{fig:design}, Step~\circled{0}) provides the seed implementation and static domain prompt. The seed defines the starting design, while the prompt provides fixed guidance on mechanisms, architectural signals, or design restrictions.

The framework's configuration inputs, marked by gear icons in Figure~\ref{fig:design}, include the LLM and evolutionary framework, simulator and hardware model, trace database, scoring function, and hardware
budgets. Hardware modeling provides cost estimates and budget checks, simulation returns performance metrics and diagnostics, and the scoring function quantifies evaluation results, reflecting the optimization objective. Training workloads provide feedback during evolution, while held-out workloads evaluate the final design.

Defined component interfaces make \arch{} modular, allowing different search methods, LLMs, simulators, and hardware models to be integrated into the same workflow. Beyond incorporating new tools and models, this modularity supports different architectural domains that require their own evaluation tools and objectives, without rebuilding the optimization loop. Once initialized and configured, these choices remain fixed throughout a run, so all candidates are evaluated under the same conditions.

\subsection{Design Evolution}

Design evolution (Figure~\ref{fig:design}, Step~\circled{1}) is performed by the evolutionary framework and LLM. At each iteration, the framework selects a parent from the design population and receives a dynamic prompt composed of its design and prior feedback. The LLM then produces an evolved design. The search remains bounded by the architect-defined design directives and requirements. The candidate is then passed to the evolutionary agent (Figure~\ref{fig:design}, Step~\circled{2}) for validation and evaluation.

\subsection{\arch{}'s Evolutionary Agent}

The evolutionary agent (Figure~\ref{fig:design}, Step~\circled{2}) sits between design evolution and microarchitectural simulation, and its role is to turn the selected evolutionary framework into a constrained architectural design search. It coordinates candidate generation, validation, simulation, and scoring, returning both scores and failures to the LLM for subsequent evolutions. Because evolved designs may be invalid, hardware-infeasible, or prohibitively expensive to evaluate, the agent enforces two design principles.

\noindent\textbf{Machine-checkable candidates.} Generated designs follow a predefined output contract that enables automatic validation and prevents non-executable candidates from consuming simulation time.

\noindent\textbf{Hardware feasibility during search.} Area and power are enforced as search constraints. As shown in Figure~\ref{fig:design}, the hardware model evaluates each candidate before simulation and rejects overprovisioned designs. Later, execution activity enables dynamic-power validation. This both prevents evolution from buying performance with additional hardware and avoids wasting compute on already-infeasible candidates.

Accepted candidates proceed to the simulator (Figure~\ref{fig:design}, Step \circled{3}), while rejected ones return negative scores and failure diagnostics, steering evolution away from invalid designs. The agent spans architectural domains, turning generic evolutionary search into architectural exploration through validity checks, hardware constraints, and simulator feedback.

\subsection{Integration with Microarchitectural Simulators}
\label{subsec:uarch}

The simulator (Figure~\ref{fig:design}, Step~\circled{3}) serves as an interchangeable execution backend behind a fixed interface. This interface separates the evolutionary search from simulator-specific details and has two roles.

\noindent\textbf{Design interface.} For each architectural domain, the simulator exposes the points where an evolved design may act and the signals it may observe. A replacement design, for example, selects victims and updates replacement state, while a prefetcher observes cache accesses and proposes requests. Evolution may change the mechanism within this boundary, but not the surrounding microarchitecture. The interface also exposes hardware activity needed for power modeling.

\noindent\textbf{Measurement interface.} The simulator executes accepted candidates on the training traces and reports performance measurements and domain-specific statistics, such as IPC, cache misses, or branch mispredictions to the evaluator. 

This abstraction makes \arch{} portable across simulators and architectural components: a new backend only needs to expose the corresponding design and measurement interfaces, without changing the evolutionary search.

\subsection{\arch{}'s Evaluator}

The evaluator (Figure~\ref{fig:design}, Step~\circled{4}) translates simulator measurements into the scalar fitness that drives evolution. Its configurable scoring function allows the architect to define the optimization objective. For example, it can optimize raw IPC, performance per watt, or domain-specific metrics.

Importantly, scoring is separated from diagnostic feedback. Metrics such as cache misses, and branch mispredictions may be reported without affecting overall ranking, helping explain design improvements or regressions. Scores and diagnostics guide subsequent evolutions, and the highest-scoring design is evaluated in the end on held-out traces.

\subsection{Prompt Design}

The prompt is the harness through which \arch{} steers the LLM during evolution. It combines a static prompt provided at initialization with a dynamic prompt constructed from feedback during the search.

\noindent\textbf{Static prompt.} As described earlier, the static prompt is written once by the architect and remains fixed throughout evolution. It communicates the design goal, the seed, and any guidance the architect wants the LLM to use. 

\noindent\textbf{Dynamic prompt.} The dynamic prompt provides live feedback from the evolutionary search back to the LLM. For each iteration, it includes the selected parent design together with its score, diagnostic metrics, and feedback from prior evaluations. For example, the LLM can learn that a candidate improved IPC but increased misses, or that a previous iteration violated a hardware constraint.

\subsection{The Role of the Architect}
\label{sec:design_human}

Several inputs and configurations remain explicit architectural decisions: the seed that initializes search, the objective it optimizes, the constraints candidates must satisfy, the workloads used for training, and the guidance provided through the prompt. \arch{} deliberately leaves these choices open because their effectiveness depends on the architectural domain and optimization goal.

This division of responsibility makes \arch{} a co-design workflow: the architect defines the search conditions and what constitutes progress, while the LLM explores designs within them at a scale impractical for manual iteration. In Section~\ref{sec:results} we evaluate the impact of several of these decisions on the resulting designs.
\section{Use Cases for Microarchitectural Design Exploration}
\label{sec:use_cases}

We instantiate \arch{} across three extensively studied microarchitectural domains: cache replacement, data prefetching, and branch prediction. For each use case, we select established designs as seeds and evolve them under specified objectives and hardware budgets. We first introduce the search strategy, then describe the evolved mechanisms and signals, and finally highlight how the resulting designs improve upon their seeds.

\subsection{Use Case 1: Evolving Cache Replacement Designs}

Cache replacement is our first use case. We seed evolution with Mockingjay~\cite{mockingjay}, a SoTA reuse-distance predictor, producing EVO-MJ, and evaluate final results relative to LRU as a baseline. We also seed evolution from LRU, producing EVO-LRU, to study seed sensitivity (Section~\ref{sec:weakseed}). We also compare against Hawkeye~\cite{hawkeye}, SHiP~\cite{ship}, and DRRIP~\cite{drrip}. Table~\ref{tab:replacement} summarizes these designs, along with the evolved ones, their configurations, and their storage requirements.

\begin{table}[!t]
\caption{Configuration of evaluated replacement designs.}
\label{tab:replacement}
\footnotesize
\setlength{\tabcolsep}{3pt}
\renewcommand{\arraystretch}{1.1}

\begin{tabularx}{\columnwidth}{
    |>{\raggedright\arraybackslash}p{0.23\columnwidth}
    |>{\raggedright\arraybackslash}X
    |>{\raggedleft\arraybackslash}p{0.13\columnwidth}|
}
\hline

\textbf{DRRIP}~\cite{drrip}
& 2-bit RRPV/line, 10-bit PSEL, 5-bit BRRIP counter
& \textbf{8.0 KiB} \\
\hline

\textbf{Hawkeye}~\cite{hawkeye}
& 3-bit RRPV/line, $2\times$2K SHCT, 2800-entry sampler,
$32\times128$ OPTgen
& \textbf{30.7 KiB} \\
\hline

\textbf{Mockingjay}~\cite{mockingjay}
& 5-bit ETR/line, 2K-entry RDP, 5-way sampled cache
& \textbf{33.1 KiB} \\
\hline

\textbf{EVO-LRU}
& 128-set hit count., LLC set state, 32-bank heat, 4K-entry PC cost, 4b line cost
& \textbf{33.4 KiB} \\
\hline

\textbf{SHiP}~\cite{ship}
& 2-bit RRPV/line, 16K-entry SHCT (3-bit), 4K-entry sampler
& \textbf{34.0 KiB} \\
\hline

\textbf{EVO-MJ}
& Mockingjay, 1-bit per-line miss-cost,
$8\mathrm{K}\times5$-bit no-reuse counters
& \textbf{42.1 KiB} \\
\hline

\end{tabularx}
\end{table}

\subsubsection{EVO-MJ: A cost and stream-aware Mockingjay.} 

\begin{figure}[t]
\centering
\includegraphics[width=\columnwidth]{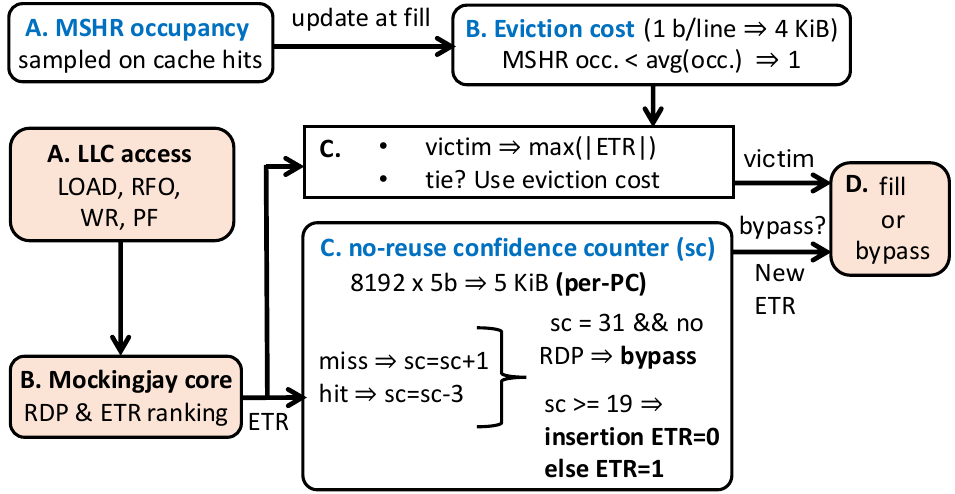}
\caption{EVO-MJ architecture. Orange components are inherited from Mockingjay. The remaining mechanisms are introduced by evolution.}
\label{fig:evomj}
\end{figure}

Figure~\ref{fig:evomj} illustrates EVO-MJ, which preserves Mockingjay's reuse-distance predictor (RDP), sampled training, and expected-time-to-reuse (ETR) ranking. It adds a per-line miss-cost bit for resolving victim ties and a per-PC no-reuse counter that supplies an ETR when the RDP has no prediction. Our analysis attributes the performance gain primarily to the latter, which improves line retention and LLC load hits; cost awareness only refines otherwise-equivalent victim choices.

\noindent\textbf{LLC access and no-reuse confidence.} As shown by the lower path in Figure~\ref{fig:evomj}, LLC accesses still use Mockingjay's RDP and ETR ranking. EVO-MJ adds a 5-bit per-PC no-reuse counter, $sc$, for LOADs and Requests For Ownership (RFOs): misses increment $sc$, while hits decrement it by three. When the RDP has no prediction, $sc\geq19$ gives $\mathrm{ETR}=0$; otherwise $\mathrm{ETR}=1$. Prefetch and WRITE handling remain unchanged. This path drives most of EVO-MJ's gain by distinguishing likely streaming from other unknown PCs, improving retention and LLC load hits.

\noindent\textbf{Victim selection.} As shown by the upper path in Figure~\ref{fig:evomj}, EVO-MJ adds a one-bit miss-cost signal per LLC line. At fill, it compares sampled MSHR occupancy with its running average: low occupancy indicates less MLP and thus a more costly future miss, so the line is marked expensive. Mockingjay's $|\mathrm{ETR}|$ ranking remains unchanged; EVO-MJ uses this bit only on ties, evicting a cheap line when possible. This preserves lines whose eviction would cause more costly misses without overriding Mockingjay's reuse prediction.

\noindent\textbf{Fill or bypass.} EVO-MJ adds one bypass case to Mockingjay: a LOAD or RFO with no RDP prediction is bypassed when its no-reuse counter saturates at $sc=31$, avoiding cache pollution from accesses with strong no-reuse evidence.

\subsubsection{EVO-LRU: Evolved cache replacement from LRU} 

\begin{figure}[t]
\centering
\includegraphics[width=0.9\columnwidth]{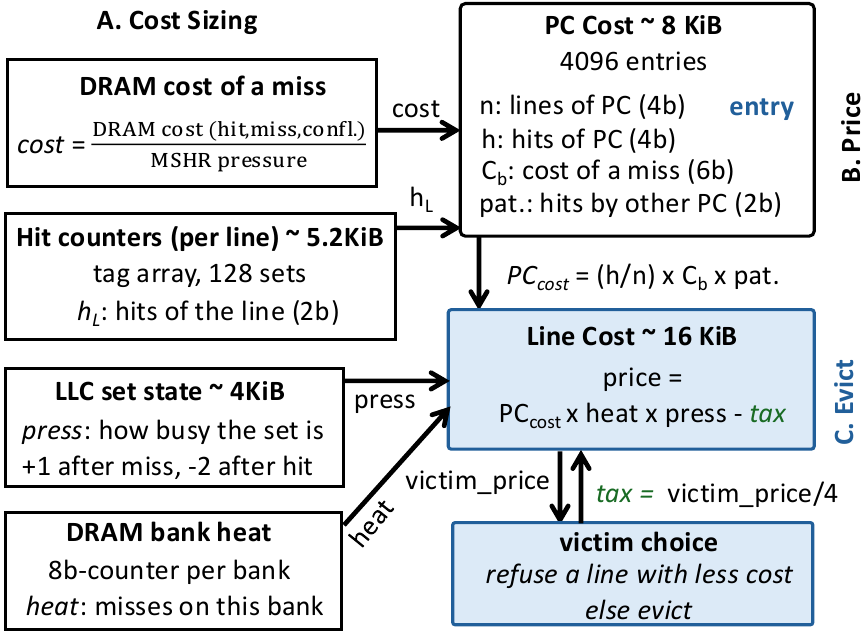}
\caption{The architecture of EVO-LRU, the evolved cache replacement design from LRU.}
\label{fig:evolru}
\end{figure}

EVO-LRU replaces LRU's recency ordering with an estimate of the cycles saved by retaining each line. Like Mockingjay, its goal is to rank each LLC line by a trained value. But instead of using Mockingjay's ETR, EVO-LRU ranks them by the expected miss cycles saved if they are kept. Figure~\ref{fig:evolru} shows how evolution constructs this signal: it first estimates a cost per PC, refines it into a per-line price, and then uses that price for admission and victim selection. In Section~\ref{sec:results} we will see how these two signals compare.

\noindent\textbf{PC-Cost sizing.} EVO-LRU learns a PC-level cost from miss cost and reuse. For each miss, DRAM row state provides a base cost, which is scaled by MSHR occupancy to estimate how much of that cost is exposed after accounting for MLP. In parallel, a sampled cache tracks a subset of LLC lines and attributes hits to their installing PC, estimating the reuse earned by that PC's lines. The PC table in Figure~\ref{fig:evolru} combines these signals with an evolved cross-PC reuse signal, which captures reuse by PCs other than the installer, to estimate the miss cycles saved by retaining a line.

\noindent\textbf{LLC-Line price.}
EVO-LRU then refines this PC-level cost using the two signals shown in Figure~\ref{fig:evolru}: LLC-set pressure and DRAM-bank heat. Set pressure captures competition for cache capacity, while bank heat is a measure of relative memory-bank traffic, increasing the value of lines mapped to banks receiving a larger share of recent misses. Their combination produces the price stored with each LLC line.

\noindent\textbf{Victim Eviction.}
EVO-LRU ranks all 16 ways by this price and selects the minimum. An arriving line that cannot beat the victim is bypassed. Otherwise it replaces the victim and pays a fraction of its price as a replacement tax. 

Thus, from an LRU seed, evolution independently converges toward a cycle-cost-based replacement design~\cite{csopt,qureshi_mlp,care}, ranking LLC lines by exposed miss cost rather than reuse distance. It further strengthens this MLP-aware ranking with two new derived signals: cross-PC reuse and relative memory-bank traffic.

\subsection{Use Case 2: Evolving Prefetching}

Data prefetching is our second use case. We seed evolution with VA/AMPM Lite~\cite{vaampm}, a compact prefetcher in the L1D cache, and report final performance relative to a no-prefetch baseline. We also seed with PC-Stride at the L2 cache, to study seed sensitivity (Section~\ref{sec:weakseed}). 

We compare with the following L1D prefetchers: IPCP~\cite{ipcp}, Berti~\cite{berti}, VA/AMPM~\cite{vaampm} and MLOP~\cite{mlop}, and the L2 prefetchers: Bingo~\cite{bingo}, SPP~\cite{spp}, Pythia~\cite{pythia}, and SPP+PPF~\cite{spp,ppf}. Table~\ref{tab:prefetchers} lists their configuration and storage needs.

\begin{table}[!t]
\caption{Configuration of evaluated prefetchers.}
\label{tab:prefetchers}
\footnotesize
\setlength{\tabcolsep}{3pt}
\renewcommand{\arraystretch}{1.1}

\begin{tabularx}{\columnwidth}{
    |>{\raggedright\arraybackslash}p{0.22\columnwidth}
    |>{\raggedright\arraybackslash}X
    |>{\raggedleft\arraybackslash}p{0.13\columnwidth}|
}
\hline

\multicolumn{3}{|l|}{\textit{L1D Prefetchers}} \\
\hline

\textbf{AMPM}~\cite{vaampm}
& 128-entry region table, 64-block access/prefetch maps, degree 2
& \textbf{3.0 KiB} \\
\hline

\textbf{Berti}~\cite{berti}
& 152-entry LT, $8\times16$ history, 64-entry BT ($\times16$ deltas)
& \textbf{5.7 KiB} \\
\hline

\textbf{MLOP}~\cite{mlop}
& 128-entry AMT, 500-update, 16-degree
& \textbf{6.5 KiB} \\
\hline

\textbf{EVO-AMPM}
& 128-entry region, 256 IP, 256 pattern, 128 NDT, 16 offsets
& \textbf{8.3 KiB} \\
\hline

\textbf{IPCP}~\cite{ipcp}
& 1K IP table, 4K DPT, 16-entry GHB
& \textbf{15.1 KiB} \\
\hline

\multicolumn{3}{|l|}{\textit{L2 Prefetchers}} \\
\hline

\textbf{SPP}~\cite{spp}
& 256-entry ST, $512\times4$-way PT, 1K filter, 8-entry GHR
& \textbf{6.8 KiB} \\
\hline

\textbf{Pythia}~\cite{pythia}
& 2 features, 3 tilings, 128 tiles, 15 actions; 64-entry ST, 256-entry PT
& \textbf{26.3 KiB} \\
\hline

\textbf{Bingo}~\cite{bingo}
& 2 KiB region, 64-entry FT, 128AT, 4K 16-way PHT, 128-entry streamer
& \textbf{34.3 KiB} \\
\hline

\textbf{EVO-Stride}
& 1K PC-stride, 256 page-stream, 256 AT, 1K PHT,
1K cost table, 1K issued, 512 shadow
& \textbf{42.3 KiB} \\
\hline

\textbf{SPP+PPF}~\cite{ppf}
& SPP w/ $2\mathrm{K}\times4$-way PT, 9-feature perceptron
(22.7K weights), $2\times$1K filters
& \textbf{71.4 KiB} \\
\hline

\end{tabularx}
\end{table}

\subsubsection{EVO-AMPM: The evolved L1D Prefetcher}

\begin{figure}[t]
\centering
\includegraphics[width=\columnwidth]{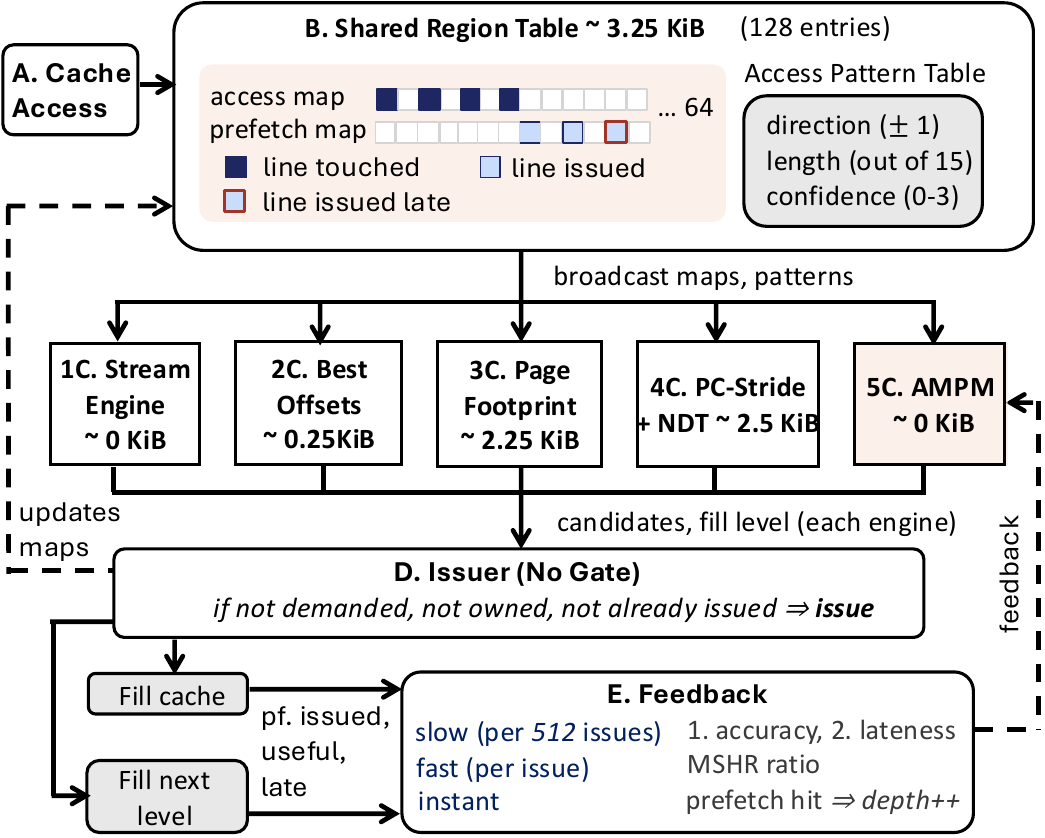}
\caption{The architecture of EVO-AMPM. The AMPM engine and bitmaps (marked with orange) are inherited from the seed. All other components are introduced by evolution.}
\label{fig:ampm_opus}
\end{figure}

EVO-AMPM evolves AMPM from a single bitmap-based predictor into a coordinated multi-engine prefetcher. As Figure~\ref{fig:ampm_opus} shows, each cache access updates shared per-page state, five complementary engines generate candidates, a common issuer removes redundant requests, and global feedback adapts their aggressiveness and depth. Thus, the gain does not come from a new prediction signal, but from combining existing predictors while coordinating their state and actions.

\noindent\textbf{Shared Region Table and engines.} EVO-AMPM retains AMPM's access and prefetch bitmaps and turns them into shared state for all five engines. Evolution augments this state with compact stream and footprint information, allowing Stream, BOP~\cite{bop}, Footprint~\cite{sms}, PC-Stride+NDT~\cite{baerchen,dcpt,vldp}, and AMPM~\cite{vaampm} to target different access patterns from the same page history. The shared bitmaps also expose whether an issued prefetch was useful or late. Compared with the AMPM seed, this broadens prediction coverage while allowing multiple engines to reuse the same per-page history.

\noindent\textbf{Global issuer.} As shown next in Figure~\ref{fig:ampm_opus}, candidates from all engines meet at a thin global issuer. It only removes candidates targeting the current demand line, duplicate candidates, and already-issued requests. EVO-AMPM therefore preserves the specialization of each engine while preventing their combination from generating redundant traffic.

\noindent\textbf{Feedback Module.} Finally, issued prefetches feed accuracy, lateness, MSHR pressure, and prefetch-hit information back to the engines. Evolution uses these signals
to adapt active engines and prefetch depth.
EVO-AMPM therefore improves over AMPM by expanding the set of recognizable access patterns while dynamically controlling the extra traffic that this broader coverage creates. Total storage is \textbf{8.3\,KiB} (Table~\ref{tab:prefetchers}).

\subsubsection{EVO-Stride: The evolved L2 Prefetcher}

\begin{figure}[t]
\centering
\includegraphics[width=\columnwidth]{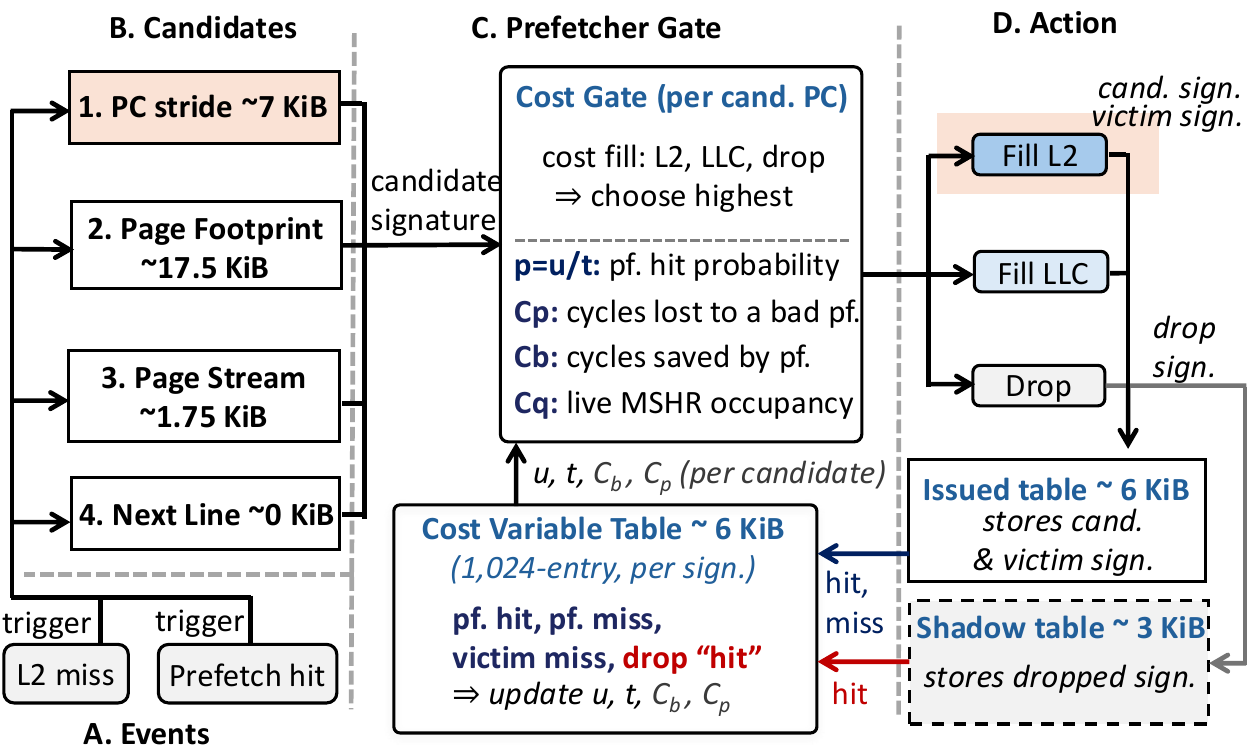}
\caption{The architecture of EVO-Stride. The PC-Stride and L2 fill only (marked with orange) are properties of the seed. The rest of the design is invented by the evolution.}
\label{fig:alfevo}
\end{figure}

EVO-Stride evolves PC-Stride into a multi-engine prefetcher, combining predictors similar to EVO-AMPM, but replacing global control with a per-candidate cost gate. As Figure~\ref{fig:alfevo} shows, four generators propose candidates, which are priced in cycles and independently filled in L2, LLC, or dropped.

\noindent\textbf{Prefetch Engines.} The retained PC-Stride engine is complemented by three more prefetching engines: Page Footprint, Page Stream, and Next Line, broadening coverage beyond repeated per-PC strides (Figure~\ref{fig:alfevo}, Step B). The engines produce candidate signatures that identify both their source and access pattern, allowing later feedback to distinguish predictions from different behaviors.

\noindent\textbf{Cost Gate.} The main evolved mechanism is the gate (Figure~\ref{fig:alfevo}, Step C). For each candidate, it learns the benefit of a successful prefetch, the cost of a harmful one, its hit probability, and current MSHR pressure. These signals are expressed in cycles and used to score three actions: fill L2, fill LLC, or drop. The highest-valued action is selected. Unlike EVO-AMPM's global feedback, EVO-Stride applies its learned cost model at candidate granularity.

\noindent\textbf{Action and feedback.} Issued and dropped candidates are tracked (Figure~\ref{fig:alfevo}, Step D) so later prefetch hits, misses, victim misses, and demands for dropped lines can update the gate's cost estimates. This feedback lets EVO-Stride issue more aggressively when the predicted cycle benefit outweighs the cost, rather than maximizing prefetch accuracy alone. The resulting gain comes largely from L2 residency: extra prefetches keep working-set lines in the L2 long enough to serve repeated demand accesses. Total storage is \textbf{42.3\,KiB}.  

\subsection{Use Case 3: Evolving Branch Prediction}

Our final use case is branch prediction, where we optimize IPC per watt, using whole-processor power to account for the pipeline effects of mispredictions. Evolution starts from 64\,KiB TAGE-SC-L~\cite{tagescl} and produces EVO-TAGE.
We compare against Bimodal, 8/64\,KiB TAGE-SC-L, 8/64\,KiB MPP~\cite{mpp}, Gshare~\cite{gshare}, and Hashed Perceptron~\cite{hashed_perceptron}. A summary of all evaluated designs is presented in Table~\ref{tab:branch_predictors}.

\begin{table}[!t]
\caption{Configuration of evaluated branch predictors.}
\label{tab:branch_predictors}
\footnotesize
\setlength{\tabcolsep}{3pt}
\renewcommand{\arraystretch}{1.1}

\begin{tabularx}{\columnwidth}{
    |>{\raggedright\arraybackslash}p{0.26\columnwidth}
    |>{\raggedright\arraybackslash}X
    |>{\raggedleft\arraybackslash}p{0.18\columnwidth}|
}
\hline

\textbf{Hashed Perceptron}~\cite{hashed_perceptron}
& 64K Weight Tables, History Registers
& \textbf{64 KiB} \\
\hline

\textbf{Gshare}~\cite{gshare}
& 4K Direction Table, 14-bit GHR
& \textbf{4 KiB} \\
\hline

\textbf{Bimodal}
& 4K Direction Table
& \textbf{4 KiB} \\
\hline

\textbf{TAGE-SC-L 8/64 KiB}~\cite{tagescl}
& 6.9K/56K TAGE, 1K/6.4K SC,
$<$0.1K/0.2K LP, --/1K Histories
& \textbf{8/64 KiB} \\
\hline

\textbf{MPP 8/64 KiB}~\cite{mpp}
& 7.97K/58.65K Weight Tables,
0.27K/1.19K Histories/State,
--/4.4K Branch Filter
& \textbf{8/64 KiB} \\
\hline

\textbf{EVO-TAGE}
& 8×1K 7-bit weight tables, 160 B local histories
& \textbf{7.17 KiB} \\
\hline

\end{tabularx}
\end{table}

\begin{figure}[t]
\centering
\includegraphics[width=\columnwidth]{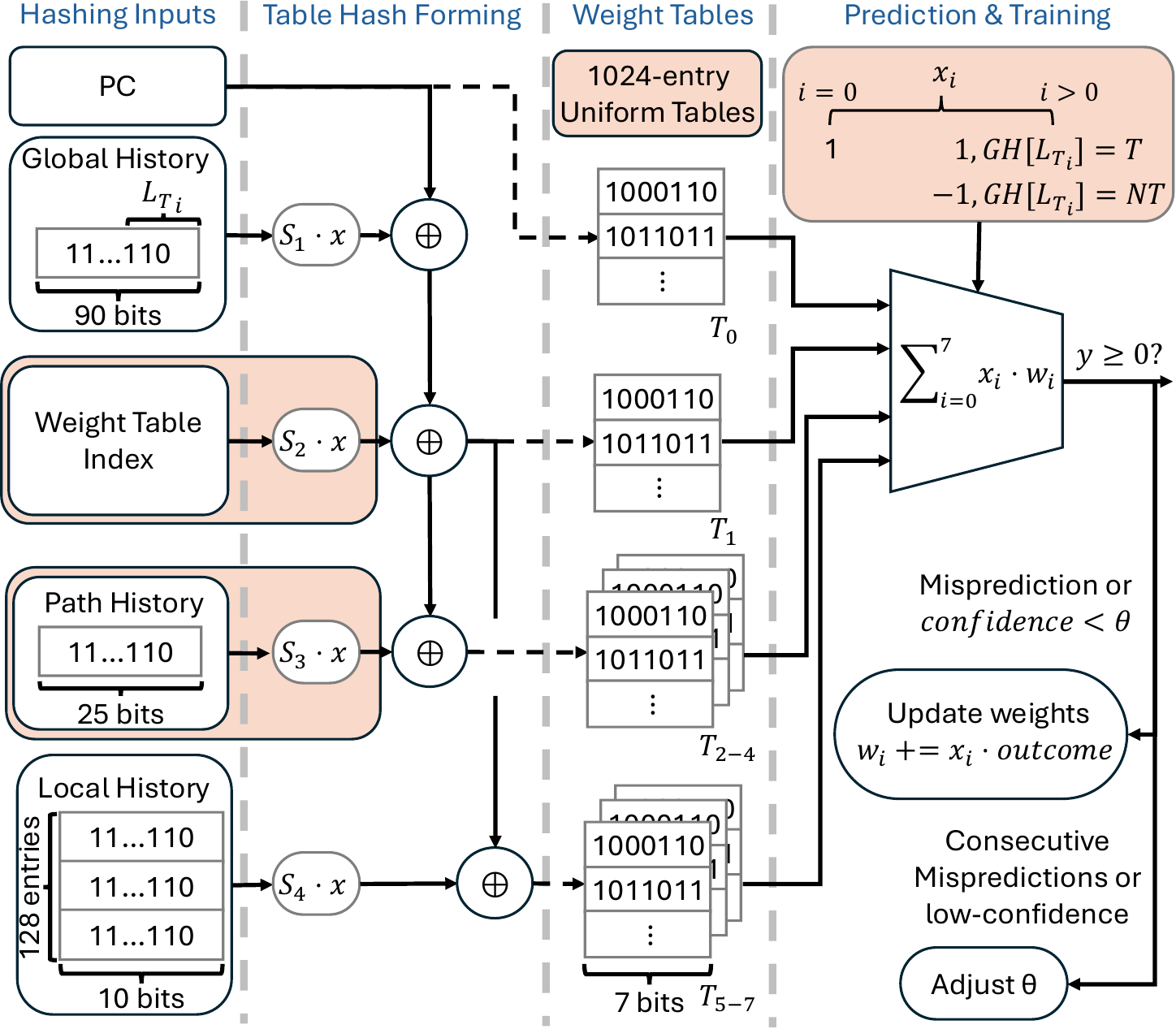}
\caption{The architecture of EVO-TAGE. The design is closer to MPP~\cite{mpp} with key differences being highlighted in orange.}
\label{fig:branch_69}
\end{figure}

\subsection{EVO-TAGE: The evolved branch predictor}
EVO-TAGE deviates form TAGE-SC-L~\cite{tagescl}, replacing its tagged prediction tables with eight, area-efficient tables of 1024 signed 7-bit weights (Figure~\ref{fig:branch_69}). For each branch, it reads one weight from each table and combines them into a sum whose sign determines whether to predict taken or not-taken.

\noindent\textbf{Selecting the weights.} The eight tables differ in the information used to select a weight. One uses only the current branch's address, providing a bias toward its usual outcome. All seven remaining tables use the address and global history: the taken/not-taken outcomes of recently executed branches, with a different history length ($L_{T_i}$) for each table. In addition, their hashes directly incorporate their index, multiplied by an odd integer salt $S_2$, a feature that is not present in MPP~\cite{mpp}. Of these seven tables, one uses no additional history, three also include a 25-bit summary of recent branch addresses (path history), and three also include the current branch's last ten outcomes (local history). Signal mixing is performed using XOR, which is also applied between individual signals and right-shifted versions of themselves. Examples of such signals are the PC and each table's final hash.

\noindent\textbf{Prediction.} The weighted sum from the tables determines the prediction: its sign indicates the branch outcome, and its magnitude indicates confidence. The address-only table contributes its weight directly. Every other table adds or subtracts its weight according to a selected global-history bit, adding for taken and subtracting for not-taken.

\noindent\textbf{Training.} The predictor updates the selected weights after a misprediction or when confidence falls below a threshold. The bias increases by one for a taken outcome and decreases by one for not-taken. Each remaining weight increases by one when the actual outcome matches its own specific global history bit, and decreases by one otherwise. These updates move the sum towards the correct outcome, following the principle used in perceptron prediction~\cite{perceptron}. The threshold is adjusted during execution aiming to balance update events caused by mispredictions and low-confidence predictions, as in MPP~\cite{mpp} and O-GEHL~\cite{ogehl}.

Unlike MPP, EVO-TAGE uses uniformly sized tables, includes global history when indexing all history-based tables, and uses global history to decide whether to add or subtract their weights. For path history, it replaces the 16-bit value array used in MPP with a 25-bit rolling hash, which is also updated by non-conditional branches save for direct jumps. Moreover, it ignores many of MPP's independently specified features such as IMLI and MODPATH. The complete predictor requires \textbf{7.17\,KiB} of state.

\subsection{What does the evolution discover?}

Across these five designs, evolution primarily discovers new mechanism combinations and signals rather than entirely novel designs. Both prefetchers combine complementary engines, while EVO-TAGE mixes well-studied branch history signals. Such combinations are non-trivial because mechanisms compete for shared state and bandwidth, and their effects are not additive: improving one can reduce another's benefit. Evolution also introduces new signals that materially improve decisions, including cross-PC reuse and relative memory-bank traffic in EVO-LRU. Thus, the novelty lies in discovering how mechanisms and signals interact under hardware constraints, rather than merely tuning parameters or adding state.
\section{Methodology \& Experimental Setup}
\label{sec:methodology}

We evaluate all three use cases using the ChampSim cycle-accurate simulator~\cite{champsim}, configured as summarized in Table~\ref{tab:config}. The memory hierarchy is kept identical across all experiments, while ROB capacity and core width follow prior evaluation setups: a 256-entry ROB, four-wide core for replacement and prefetching~\cite{crc2,dpc4,mockingjay,pythia}, and a 352-entry ROB, six-wide core for branch prediction~\cite{hashed_perceptron,branchnet}.

\begin{table}[!t]
\caption{Simulated ChampSim configuration.}
\label{tab:config}
\footnotesize
\setlength{\tabcolsep}{3pt}
\renewcommand{\arraystretch}{1.1}

\begin{tabularx}{\columnwidth}{
    |>{\raggedright\arraybackslash}p{0.10\columnwidth}
    |>{\raggedright\arraybackslash}X|
}
\hline

\textbf{Core}
&
\textbf{Cache Replacement \& Data Prefetching:}
1--4 cores, 4-wide OoO @ 4\,GHz, 256-entry ROB,
64/42-entry LQ/SQ.

\textbf{Branch Prediction:}
1 core, 6-wide OoO @ 4\,GHz, 352-entry ROB,
128/72-entry LQ/SQ.
\\
\hline

\textbf{L1/L2 Cache}
&
Private, 32\,KiB (4-way) / 256\,KiB (8-way),
64\,B line, LRU, 16/32 MSHRs,
1-cycle / 10-cycle round-trip latency.
\\
\hline

\textbf{LLC}
&
2\,MiB/core, 16-way, 64\,B line,
RQ/WQ/PQ/MSHR = 32/32/32/64 (1C),
128/128/128/256 (4C),
24-cycle round-trip latency.
\\
\hline

\textbf{Main Mem.}
&
\textbf{1C}: 1 channel, 1 rank/channel;
\textbf{4C}: 2 channels, 2 ranks/channel;
32 banks/rank, 3200\,MT/s,
64-bit data bus per channel,
$t_{\mathrm{RCD}}, t_{\mathrm{RP}}, t_{\mathrm{CAS}} = 15$\,ns,
$t_{\mathrm{RAS}} = 32.5$\,ns.
\\
\hline

\end{tabularx}
\end{table}

\subsection{Benchmark Traces}

We evaluate replacement and prefetching on 213 traces from SPEC CPU2006, SPEC  CPU2017, PARSEC, Ligra~\cite{spec2006,spec2017,parsec,ligra}, and Google Workload Traces v2~\cite{google}. For replacement and prefetching, all traces have LLC MPKI $\geq 3$ on a single-core, no-prefetching, LRU baseline. Branch prediction uses 191 traces, each with branch MPKI $\geq 3$.
Trace details are summarized in Table~\ref{tab:workloads}.

Single-core experiments use 200M warmup and 1B measured instructions, while four-core experiments use 50M warmup and 200M measured instructions per core. Four-core mixes are either homogeneous, with four copies of the same trace, or heterogeneous, with four distinct traces randomly selected from the non-Google traces. We evaluate 100 such heterogeneous mixes. We report single-core performance as geometric-mean IPC speedup and four-core performance as geometric-mean weighted speedup across mixes. Traces that reach their end are replayed from the beginning~\cite{mockingjay,pythia}. We evaluate branch prediction single-core only, since its effects are primarily within the core~\cite{hashed_perceptron,tage}.

\begin{table}
  \caption{Evaluated Workload Suites.}
  \label{tab:workloads}
  \centering
  \small
  \setlength{\tabcolsep}{3pt}
  \begin{tabular}{lcccc l}
    \toprule
    \multirow{2}{*}{Suite} &
    \multicolumn{2}{c}{\#Workloads} &
    \multicolumn{2}{c}{\#Traces} &
    \multirow{2}{*}{Example Workloads} \\
    \cmidrule(lr){2-3}\cmidrule(lr){4-5}
    & Repl/Pf & Br & Repl/Pf & Br & \\
    \midrule
    SPEC06 & 15 & 10 & 40 & 22 & gcc, mcf, bwaves, \ldots \\
    SPEC17 & 12 & 8 & 39 & 20 & gcc, mcf, fotonik3d, \ldots \\
    PARSEC & 5  & 7 & 17 & 25 & canneal, raytrace, \ldots \\
    Ligra   & 13 & 13 & 72 & 80 & BFS, BellmanFord, \ldots \\
    Google   & 12 & 12 & 45 & 44 & arizona, delta, tahoe, \ldots \\
    \midrule
    Total   & 57 & 50 & 213 & 191 & \\
    \bottomrule
  \end{tabular}
\end{table}

\subsection{Evolution Setup} 

We use OpenEvolve~\cite{openevolve} as the evolutionary backend. Each run evaluates 100 candidates over 100 iterations. To limit search cost, evolution uses only the single-core configuration with 50M warmup and 200M measured instructions on random subsets of 58 and 64 traces drawn from SPEC06, SPEC17, PARSEC, and Ligra, chosen to span a broad range of LLC MPKI and branch MPKI. Google traces are excluded entirely from evolution, providing an unseen workload set for evaluating generalization. All reported results use the full benchmark sets and the evaluation methodology above.

\subsection{Hardware Cost Model}

Each candidate declares the capacity, geometry, and organization of its stateful structures, which we model with CACTI 7.0~\cite{cacti} for area, per-access read/write energy, and leakage power. During simulation, the design reports per-structure read and write counts. Multiplying these counts by their access energies and dividing by simulated execution cycles gives the mechanism's average dynamic power. Adding leakage gives its total power. We follow the same methodology to model the processor's caches and TLB.

For the rest of the processor, McPAT~\cite{mcpat} estimates peak dynamic power, scaled by IPC relative to maximum retire width to approximate average activity, with leakage added separately. Together, these estimates capture both the mechanism's direct hardware cost and its effect on processor activity. CACTI and McPAT use a 65\,nm technology node. 

Unless stated otherwise, area and power budgets are relative to the seed. Our main experiments reject candidates exceeding 30\% overhead in either metric.

\subsection{LLM, Prompts and Scoring} 

\begin{figure}[t]
\textbf{Prompt template}
\vspace{2pt}
\begin{lstlisting}[
  basicstyle=\ttfamily\fontsize{7}{8.5}\selectfont,
  frame=single,
  columns=fullflexible,
  keepspaces=true
]
You are an expert computer architect specializing in LLC
replacement policies. Evolve the provided policy [..].

# GOAL
Maximize `combined_score` = geometric mean of IPC across 58
traces, weighting every trace's RELATIVE improvement equally.
LLC miss counts are diagnostic and do NOT affect the score.
Prefer trading raw misses for hits on critical lines. [...]

# BUDGET
Area and power overhead must each stay within 30%.

# TASK
Respond with complete openevolve.{h,cc} in the following format:
// [START OF HEADER FILE: openevolve.h]
...
// [END OF HEADER FILE: openevolve.h]

// [START OF SOURCE FILE: openevolve.cc]
    #include "openevolve.h"
    // ... implementations ...
// [END OF SOURCE FILE: openevolve.cc]

# CACTI CONFIG
At the end of your response, declare each stateful structure your
branch predictor adds (prediction tables, tag tables, history
buffers, loop tables, statistical corrector tables, counters) in
exactly this format:...

# ACCESS COUNTERS
Every structure declared in the CACTI block must maintain read/write
counters, printed in replacement_final_stats() - one line per
structure, name matching the CACTI_STRUCTURE name exactly
, format exactly: ...

# TARGET SYSTEM
- LLC 2 MiB 16-way, 24-cyc, 64 MSHRs | L2 256 KiB | L1D 32 KiB
- 4 GHz, 256-ROB | DRAM 3200 MT/s, 15/45 ns row hit/conflict

# SEARCH STRATEGY
Declare your move on line 1 of the header, then execute it:
 // MOVE: NOVEL   -- progress stalled; invent a mechanism
    matching no published policy, on unused signals.
 // MOVE: IMPROVE -- still gaining; tune tables or graft
    one complementary mechanism.
\end{lstlisting}
\caption{Simplified system prompt for the cache replacement use case,
specifying the objective, output contract, target system, available signals,
feasibility budgets, and search directive.}
\label{fig:prompt_examples}
\end{figure}

\begin{figure}[t]
\vspace{2pt}
\begin{lstlisting}[
  basicstyle=\ttfamily\fontsize{7}{8.5}\selectfont,
  frame=single,
  keepspaces=true
]
# GOAL
...Since the branch prediction seed is particularly
strong and achieves very high prediction accuracy, your
focus should be on reducing the design's area and power
while maintaining accurate branch predictions.
\end{lstlisting}
\caption{System prompt goal modification for branch prediction.}
\label{fig:branch_pred_goal}
\end{figure}

We use Claude Opus 5~\cite{claude_opus} as the LLM in this instance of \arch{}.
A static system prompt explicitly defines the seed policy, optimization objective, and steering directives including the area and power budgets. It also specifies a strict LLM output contract which has to be verified by the evolutionary agent for the design to proceed with the evaluation. Finally, it provides the LLM with the specifications of the evaluated system as well as potential exploration options.
A cache replacement evolution example is presented in Figure~\ref{fig:prompt_examples} while the goal shift towards efficiency for the branch prediction use case is presented in Figure~\ref{fig:branch_pred_goal}.
During evolution, cache replacement and prefetching candidates are scored by geometric-mean IPC. For branch prediction, we score with IPC per watt, demonstrating that the framework can target a different objective.
\section{Evaluation}
\label{sec:results}

We evaluate \arch{} across cache replacement, data prefetching, and branch prediction, comparing evolved designs with their seeds and SoTA designs. We ask whether evolution can improve mature designs, discover competitive designs from weaker seeds, satisfy hardware budgets, and generalize to unseen workloads and longer simulations.

\subsection{Cache Replacement}

\begin{figure}[t]
\centering
\includegraphics[width=\columnwidth]{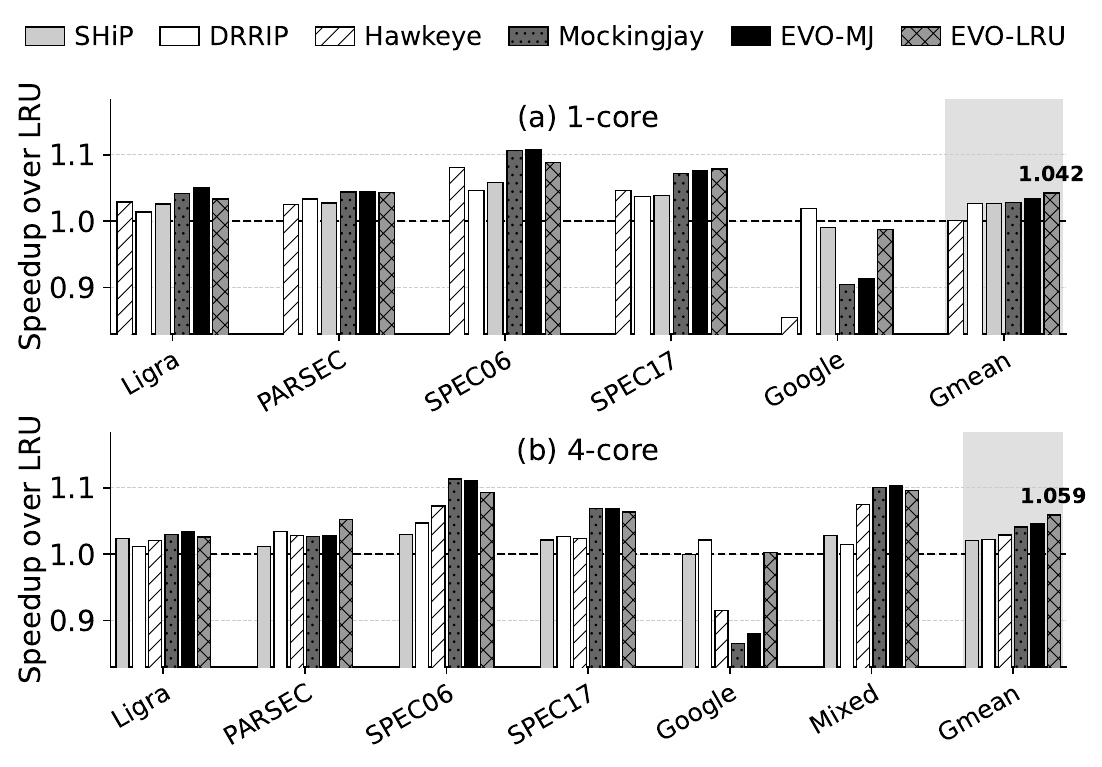}
\caption{Geometric mean speedup over LRU by benchmark suite for (a) one core and (b) four cores. \emph{Mixed} denotes heterogeneous four-application mixes.}
\label{fig:cache_speedup}
\end{figure}

We seed evolution with Mockingjay, producing EVO-MJ. Figure~\ref{fig:cache_speedup} reports geometric mean speedup over LRU for one and four cores. EVO-LRU is discussed separately in Section~\ref{sec:weakseed}.

EVO-MJ reaches 1.034$\times$ and 1.045$\times$ speedup over LRU, improving Mockingjay, the strongest baseline, by 0.55\% and 0.44\%, respectively. Its largest gain over Mockingjay occurs on Ligra. Both designs degrade sharply on the unseen Google traces, reflecting behavior inherited from Mockingjay, although EVO-MJ recovers about 1.2\% over its seed.

EVO-MJ wins primarily by correcting accesses for which Mockingjay's RDP has no prediction. Mockingjay assigns such demand lines $\mathrm{ETR}=0$, allowing streaming accesses to displace reusable data. EVO-MJ's no-reuse counter identifies these streams earlier and bypasses them once confidence saturates, preserving reusable lines. Its miss-cost signal further improves equal-ETR decisions by retaining lines whose future misses would expose more latency. This explains its larger gains when streaming and reusable accesses coexist, as in Ligra, while leaving little room to help workloads with almost no reusable LLC data.

\subsection{Data Prefetching}

\begin{figure}[t]
\centering
\includegraphics[width=\columnwidth]{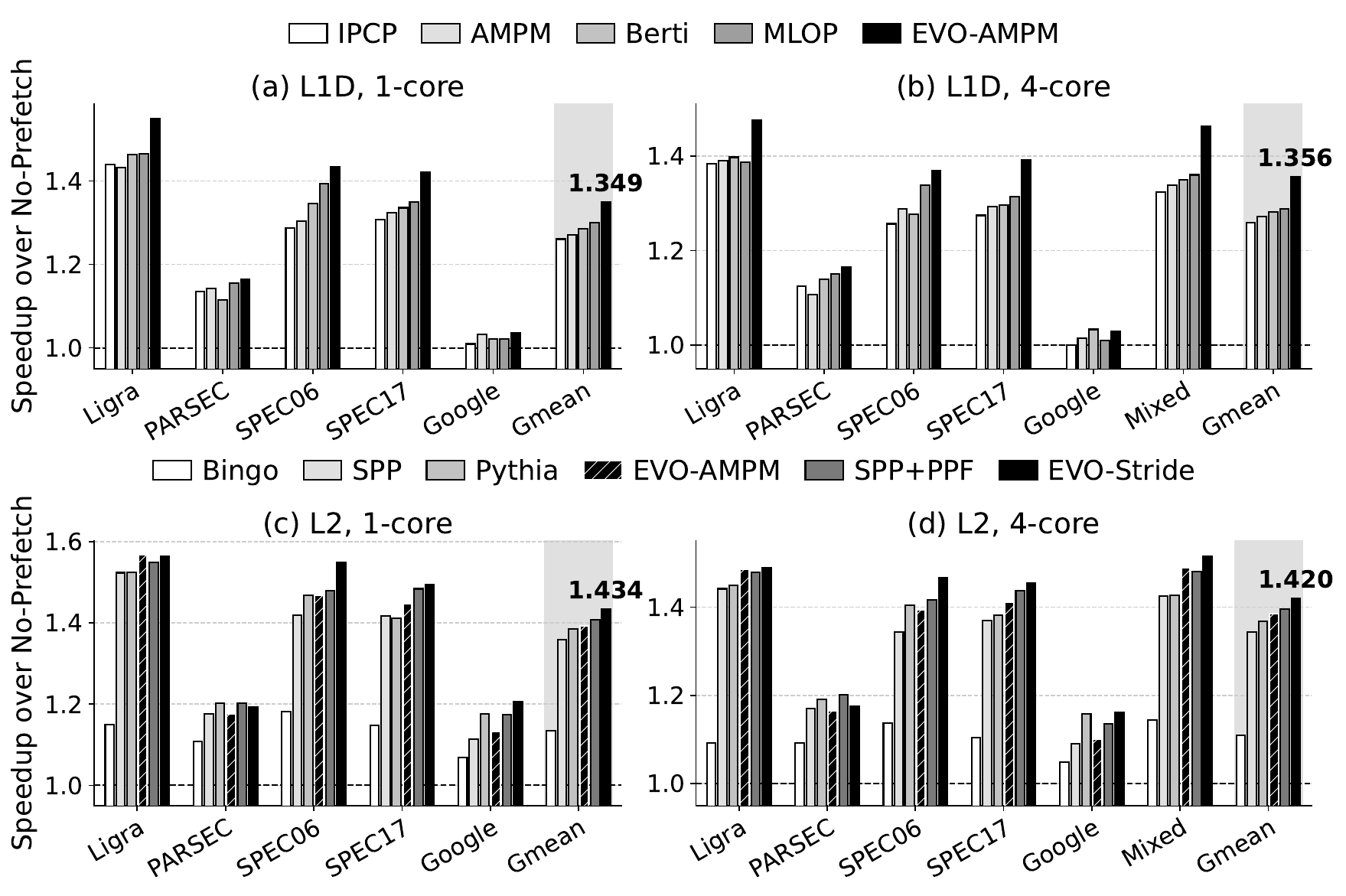}
\caption{Geometric mean speedup over no prefetching by benchmark suite for
L1D at (a) one core and (b) four cores, and L2 at (c) one core and
(d) four cores. \emph{Mixed} denotes heterogeneous four-application mixes.}
\label{fig:pref_res}
\end{figure}

For prefetching, we seed evolution with AMPM at L1D, producing EVO-AMPM. Figure~\ref{fig:pref_res}(a,b) compares EVO-AMPM with SoTA designs against a no-prefetching baseline.

On our traces, EVO-AMPM is the best evaluated L1D prefetcher, reaching $1.349\times$ and $1.356\times$ speedup on one and four cores, respectively, corresponding to an additional 4.9\% and 6.7\% speedup over MLOP and 7.8\% and 8.4\% over its AMPM seed. Its largest margins over MLOP appear on Ligra and heterogeneous mixes, adding 8.9\% and 10.3\%, respectively. Its advantage also persists under contention at four cores, even as baseline rankings change across workloads, indicating robust behavior across different access patterns.

EVO-AMPM's multi-engine design also remains well targeted. At four cores, it generates an average of 1.37B useful prefetches versus MLOP's 1.23B, with 40.7\% accuracy versus 35.9\% and 90.15\% L1D load-hit rate versus 87.53\%. Its shared state increases coverage, while the global issuer and adaptive feedback reject redundant requests and control aggressiveness, increasing coverage without sacrificing precision.

\subsection{Evolution from a Weaker Seed}
\label{sec:weakseed}

We next ask whether \arch{} can discover competitive designs when starting from a weaker seed.

\subsubsection{Cache Replacement: Evolving from LRU}

We seed evolution with LRU, producing EVO-LRU, and allow an area budget 30\% above Mockingjay's to encourage exploration. Because initial searches repeatedly converged to Mockingjay's ETR, we explicitly prohibit ETR in the prompt.

Figure~\ref{fig:cache_speedup} shows EVO-LRU's geometric mean speedup alongside EVO-MJ. EVO-LRU achieves the highest geometric mean, reaching $1.042\times$ on one core and $1.059\times$ on four, improving on Mockingjay by approximately 1.4\% and 1.7\%. Its advantage comes from better balance across workload sources. On Google workloads, Mockingjay degrades by 9.5\% and 13.5\% on one and four cores, while EVO-LRU limits the one-core loss to 1.3\% and improves four-core performance by 0.2\%, recovering 8.2\% and 13.7\% respectively. Excluding the Google traces, EVO-LRU reaches $1.057\times$ and $1.069\times$, versus Mockingjay's $1.064\times$ and $1.073\times$, trading a small amount of peak performance for better balance, yielding a robust design.

\subsubsection{Data Prefetching: Evolving from PC-Stride}

We also evolve from PC-Stride at L2, allowing an area budget 30\% above Pythia's. The search produces EVO-Stride, evaluated against SoTA L2 prefetchers and EVO-AMPM placed at L2.

Figure~\ref{fig:pref_res}(c,d) shows that EVO-Stride is the strongest L2 design overall, reaching $1.434\times$ and $1.420\times$ speedup on one and four cores. It exceeds SPP+PPF, the strongest published baseline overall, by 2.6\% and 2.0\%, respectively. Its largest gains appear on SPEC06, where it leads SPP+PPF by 6.8\% on one core and 5.0\% on four. On the unseen Google traces, it also improves over SPP+PPF by 3.2\% and 2.6\%.

EVO-AMPM also transfers well across cache levels, ranking third overall despite being evolved only at L1D, without exposure to L2 access streams, latencies, or bandwidth conditions. Together with EVO-LRU, these results highlight a seed tradeoff: mature seeds provide a robust foundation and favor refinement, whereas simpler seeds leave more room for broader exploration and can produce competitive designs.

\subsection{Branch Prediction}
\label{sub:branch_prediction}

\begin{figure}[!t]
\centering
\includegraphics[width=\columnwidth]{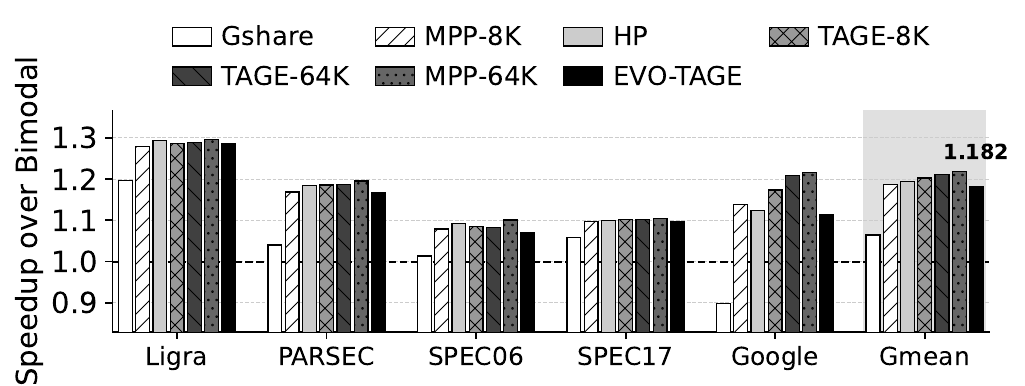}
\caption{Geometric mean speedup over bimodal by benchmark suite.}
\label{fig:branch_per_suite}
\end{figure}

\begin{figure}[t]
\centering
\includegraphics[width=\columnwidth]{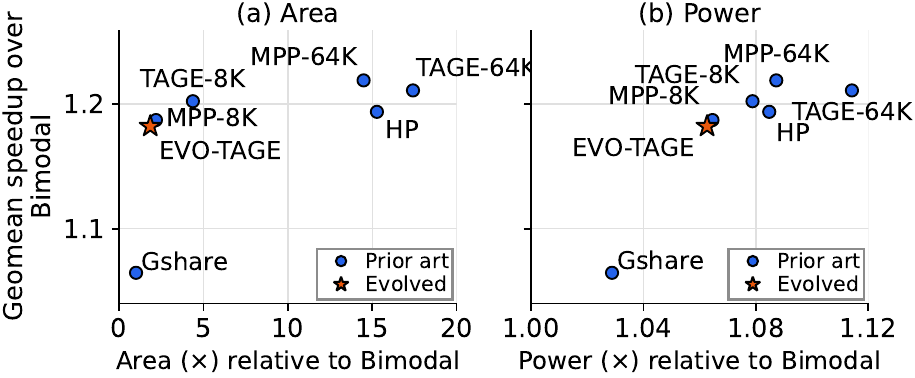}
\caption{Geometric mean IPC speedup vs.\ (a) CACTI-modeled predictor area, (b) CACTI+McPAT modeled processor power, normalized to Bimodal's corresponding values.}
\label{fig:branchevo}
\end{figure}

We seed evolution with TAGE-SC-L and optimize cost efficiency rather than raw speedup, producing EVO-TAGE.
Figure~\ref{fig:branch_per_suite} illustrates the predictors' geometric mean IPC across the five studied suites while,
Figure~\ref{fig:branchevo} plots geometric mean IPC against modeled predictor area and power, normalized to Bimodal.
EVO-TAGE retains 97.62\% and 96.98\% geometric mean IPC of its seed and the best-performing 64\,KiB MPP respectively, while taking up only 10.58\% and 12.72\% of their area and reducing processor power by 4.62\% and 2.26\%.
Except for Ligra, its performance does not match that of its closest area-wise mechanism, the 8 KiB MPP variant, trailing by 0.43\%. Still, it occupies 15.48\% less area and consumes 0.18\% less power across all workloads.
Despite ranking sixth in terms of performance, it remains on the IPC vs. Area and IPC vs. Power Pareto frontiers, and achieves the evolutionary objective by being $9.228\times$ more area efficient than its seed ($IPC/mm^2$) and $1.024\times$ more power efficient ($IPC/W$).

Overall, although seeded with TAGE-SC-L, the framework was able to recognize the area inefficiency of using large tagged tables and steer its evolved architectural policy towards a simpler perceptron design. Despite its small state size and lack of complex signals compared to MPP, it was able to achieve comparable performance across all workload suites at a fraction of the SoTA mechanism's area.

\subsection{Area Consumption of the Evolved Designs}

\begin{figure}[t]
\centering
\includegraphics[width=\columnwidth]{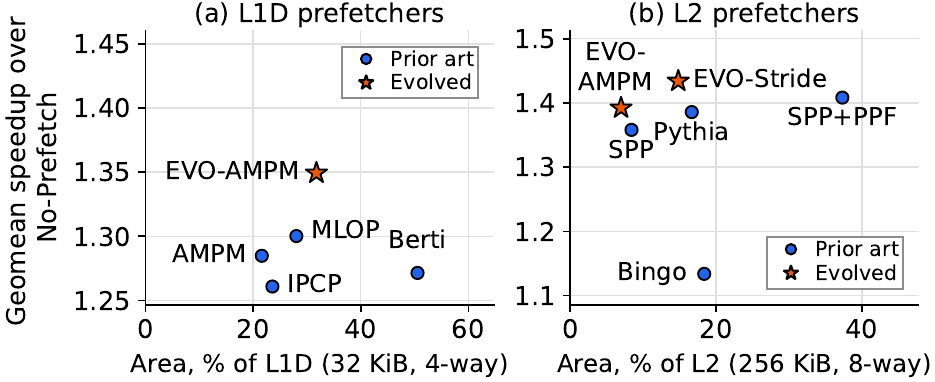}
\caption{Single-core IPC speedup vs. CACTI-modeled chip area, normalized to the area of the cache served, for (a) L1D and (b) L2 prefetchers.}
\label{fig:pareto_pref}
\end{figure}

\begin{figure}[t]
\centering
\includegraphics[width=0.9\columnwidth]{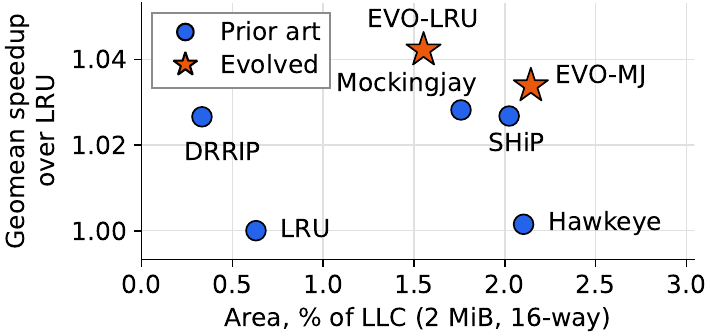}
\caption{Single-core IPC speedup vs. CACTI-modeled chip area, normalized to LLC area, for the replacement designs.}
\label{fig:pareto_repl}
\end{figure}

Next, we evaluate the hardware cost of the evolved mechanisms. Figures~\ref{fig:pareto_pref} and~\ref{fig:pareto_repl} show single-core geometric mean speedup against CACTI-modeled area, expressed as a percentage of the cache each design serves.

At L1D, EVO-AMPM is the fastest design while occupying 31.8\% of area, only 3.7\% more than MLOP and 18.8\% less than Berti. At L2, EVO-Stride is the fastest design while being 1.8\% and 22.5\% smaller than Pythia and SPP+PPF, respectively. EVO-AMPM offers a favorable tradeoff: SPP+PPF outperforms it by
1.6\% while occupying 30.4\% more L2 area.

Cache replacement tells a similar story, with each evaluated design occupying less than 3\% of LLC area. EVO-MJ consumes 2.1\%, 0.4\% more than Mockingjay, while EVO-LRU consumes 1.6\% and achieves the highest
overall speedup. Across both use cases, \arch{} produces competitive designs respecting their hardware constraints.

\subsection{Generalization to unseen workloads}

\begin{figure}[t]
\centering
\includegraphics[width=\columnwidth]{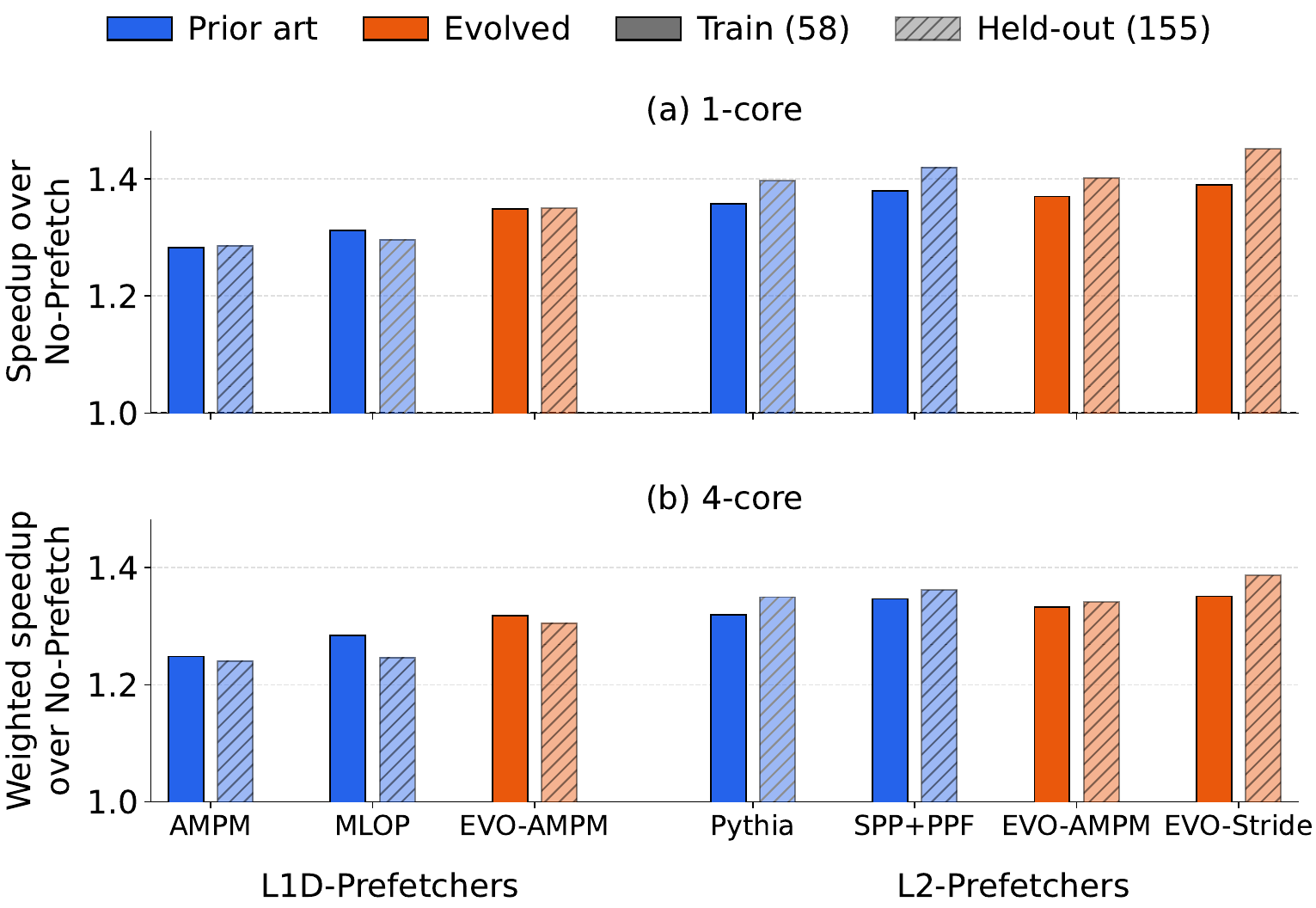}
\caption{(a) Single-core geometric mean IPC speedup and (b) Four-core geometric mean weighted IPC speedup on 58 training traces vs. 155 held-out traces, for L1D-placed and L2-placed prefetchers.}
\label{fig:train_heldout_pref}
\end{figure}

\begin{figure}[t]
\centering
\includegraphics[width=\columnwidth]{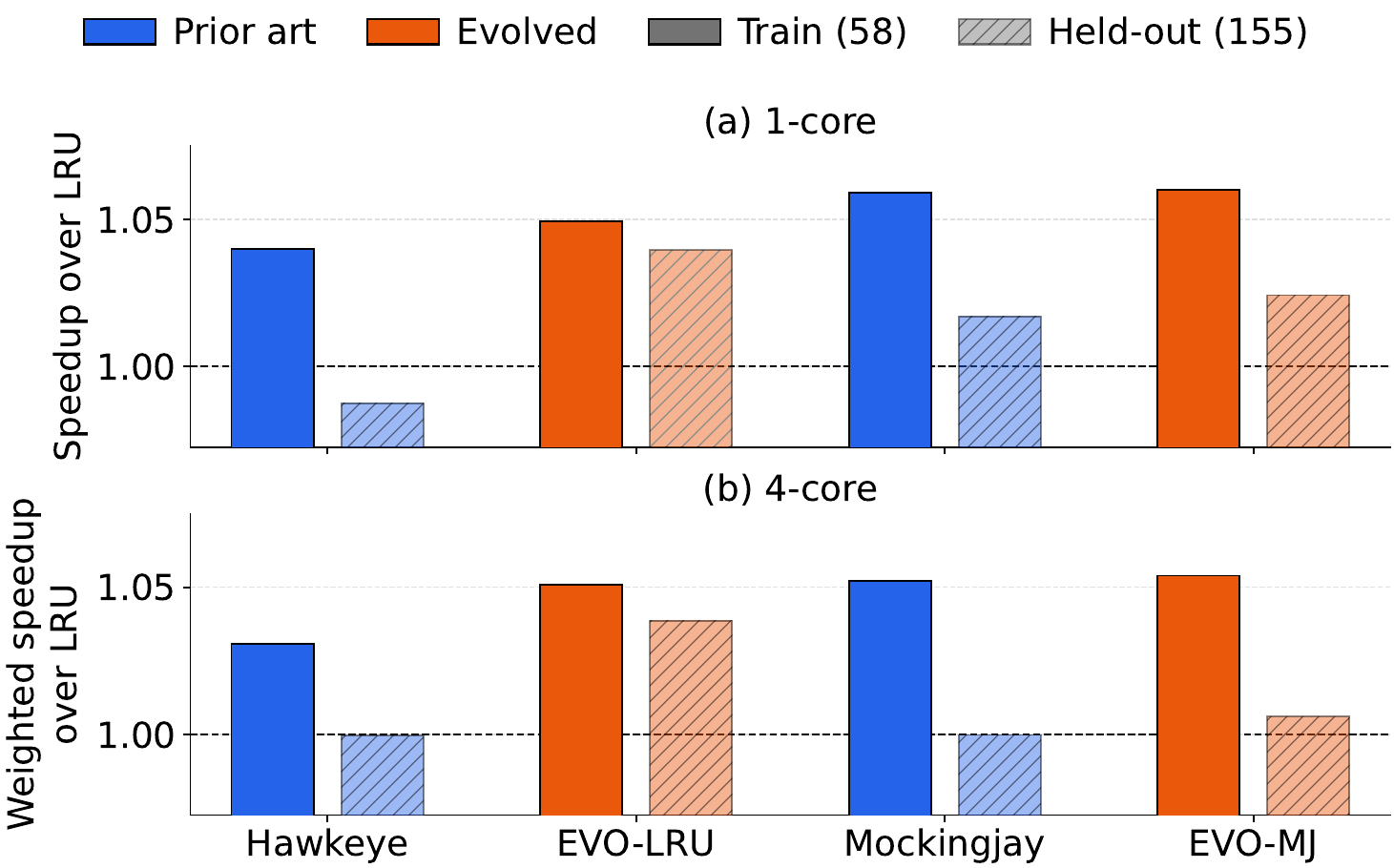}
\caption{(a) Single-core geometric mean IPC speedup and (b) Four-core geometric mean weighted IPC speedup on 58 training vs. 155 held-out traces, for replacement designs.}
\label{fig:train_heldout_cache}
\end{figure}

\begin{figure}[t]
\centering
\includegraphics[width=\columnwidth]{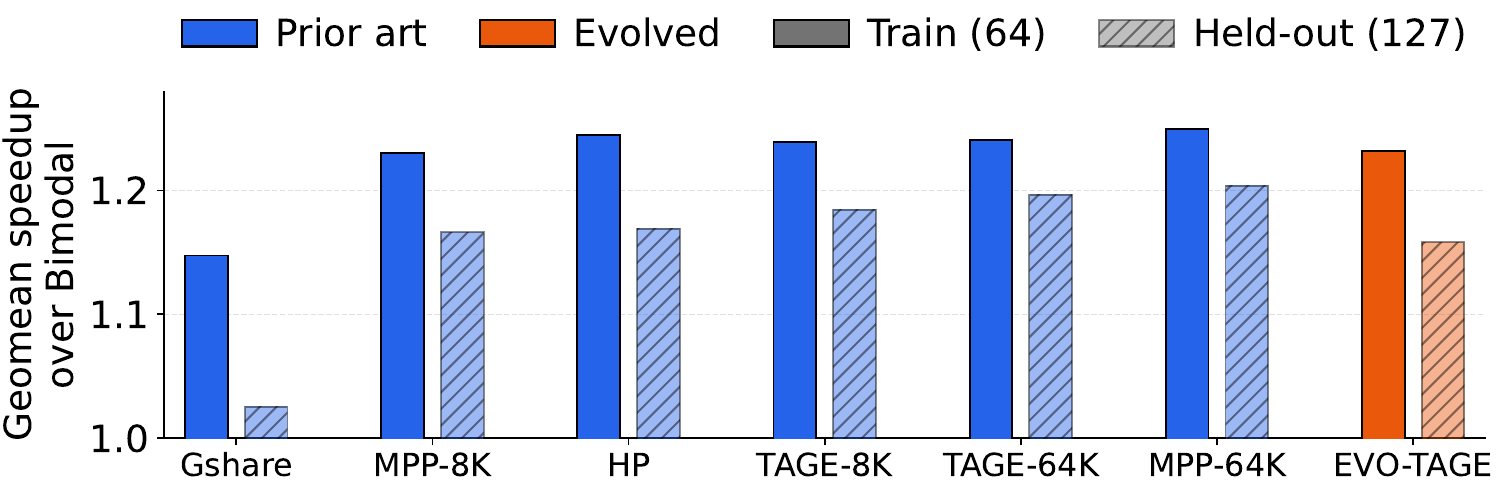}
\caption{Geometric mean IPC speedup on 64 training traces vs. 127 held-out traces, for branch predictors.}
\label{fig:train_heldout_branch}
\end{figure}

Because evolution is rewarded on 58 or 64 training traces, depending on the use case, we test if its gains generalize to 155 and 127 held-out traces, including the unseen Google workloads. Figures~\ref{fig:train_heldout_pref}, \ref{fig:train_heldout_cache}, and~\ref{fig:train_heldout_branch} report geometric mean speedup for prefetching, replacement and branch prediction respectively. 
For prefetching and replacement the four-core geometric mean weighted speedup is also reported.

Prefetcher rankings generalize well: EVO-AMPM remains the strongest L1D design and EVO-Stride the strongest L2 design on held-out traces, both for one and four cores. Their margins over the strongest prior-art designs widen rather than shrink.

Replacement designs also generalize well, with one notable shift. On training traces, EVO-MJ reaches $1.060\times$ at one core, essentially matching Mockingjay's $1.059\times$, while EVO-LRU reaches $1.049\times$. On held-out traces, EVO-LRU is best at $1.040\times$, ahead of EVO-MJ ($1.024\times$) and Mockingjay ($1.017\times$), consistent with its better balance on the unseen Google traces. Similarly, at four cores, EVO-LRU remains best on held-out traces at $1.039\times$, while EVO-MJ slightly exceeds Mockingjay ($1.006\times$ vs.\ $1.000\times$).

The same generalization trend holds for branch prediction.
The only exception is EVO-TAGE outperforming the 8 KiB variant of MPP on the training set by a marginal 0.12\% but being 0.7\% slower on the larger hold-out set.

The gains also persist over longer simulations. Evolution uses 50M warmup and 200M measured instructions per trace, whereas final evaluation uses 200M warmup and 1B measured instructions. This also motivated excluding traces whose short-window behavior could bias the search. For example, we drop \texttt{607.cactuBSSN\_s}
(SPEC17), whose replacement sensitivity is specific to the short evolution window and disappears in the longer final evaluation. This emphasizes the importance of a well structured training set including diverse and well behaved traces.
\section{Related Work}
\label{sec:related}

\noindent\textbf{LLM-driven evolutionary search.} ELM~\cite{elm} first used LLMs as mutation operators within an evolutionary loop, and FunSearch~\cite{funsearch} showed that this approach can produce mathematical advances. AlphaEvolve~\cite{alphaevolve} extended the paradigm to general code optimization, motivating a family of open-source frameworks~\cite{openevolve,openelm,adaevolve, codeevolve,shinkaevolve,deepevolve,thetaevolve}. These frameworks explore different strategies for sample efficiency, mutation diversity, and search adaptation. We build on OpenEvolve~\cite{openevolve} and apply this to CPU microarchitecture, where effective search requires hardware budgets, representative training workloads, and support for different architectural domains, objectives, and evaluation tools.

\noindent\textbf{Machine Learning for systems.} Prior work applies learning and automated search to index structures~\cite{kraska2018case}, query optimization~\cite{neo}, and program autotuning~\cite{opentuner}. ADRS~\cite{adrs} uses LLM-driven evolution to discover algorithms for cloud scheduling, LLM inference, and transaction management. Concurrent work ArchAgent~\cite{archagent} explores cache replacement through AlphaEvolve, while  ArchAgent v2~\cite{archagentv2} provides a framework for evolving only multilevel prefetchers using staged search and hardware-size feedback. Sankaralingam~\cite{alphazero_moment} proposes an automated ``Idea Factory'' for architecture research. Our work is the first to propose an end-to-end, domain-independent, open-source framework that makes the seed, optimization objective, hardware budget, and training workload explicit architect-controlled dimensions of the search. We study these choices across cache replacement, prefetching, and branch prediction, including their effects on search behavior, hardware feasibility, and generalization to unseen workloads, and examine the designs and mechanisms toward which evolution converges.
\section{Conclusion}
\label{sec:conclusion}

We introduce \arch{}, an agentic CPU microarchitecture optimization framework combining LLM-driven evolution with cycle-accurate simulation and hardware cost modeling. Across prefetching, cache replacement, and branch prediction, \arch{} matches or improves prior art. Its evolved L1D and L2 prefetchers exceed MLOP by an additional 4.9\%/6.7\% and SPP+PPF by 2.6\%/2.0\% on one/four cores. The best evolved replacement design improves Mockingjay by 1.4\%/1.7\%, while the branch predictor retains 97.62\% of TAGE-SC-L's geometric mean IPC at 10.58\% of its area.

Our results support agentic co-design, where LLM-driven search accelerates exploration within architect-defined constraints. Seed choice shapes the search trajectory, objectives and hardware budgets define tradeoffs, and training workloads influence generalization. \arch{} makes these choices explicit and configurable, enabling practical agentic exploration of microarchitectural designs. The framework will be open-sourced upon publication.

% use the ACM bibliography style
\bibliographystyle{ACM-Reference-Format}
\bibliography{references}

\end{document}